\documentclass{article}

\usepackage{microtype}
\usepackage{graphicx}
\usepackage{subfigure}
\usepackage{booktabs} 

\usepackage{hyperref}

\usepackage[accepted]{icml2023}

\usepackage{amsmath}
\usepackage{amssymb}
\usepackage{mathtools}
\usepackage{amsthm}

\usepackage[capitalize,noabbrev]{cleveref}

\usepackage{graphicx}
\usepackage{booktabs}
\usepackage{amsmath}
\usepackage{amsfonts}
\usepackage{paralist}
\usepackage{algorithm}
\usepackage{mathrsfs}
\usepackage{enumitem}
\usepackage{tabu}
\usepackage{multirow}
\usepackage{amsmath}
\usepackage{amssymb}
\usepackage{amsthm}
\usepackage{amsfonts}
\usepackage{colortbl}

\newcolumntype{H}{@{}>{\lrbox0}l<{\endlrbox}}
\newcolumntype{Z}{>{\setbox0=\hbox\bgroup}c<{\egroup}@{\hspace*{-\tabcolsep}}}

\newcommand\given[1][]{\:#1\vert\:}

\theoremstyle{plain}

\theoremstyle{definition}

\theoremstyle{remark}

\newcommand{\ours}{GeLaTo}{}

\usepackage[textsize=tiny]{todonotes}

\icmltitlerunning{Tractable Control for Autoregressive Language Generation}

\begin{document}
\twocolumn[

\icmltitle{Tractable Control for Autoregressive Language Generation}



\icmlsetsymbol{equal}{*}

\begin{icmlauthorlist}
\icmlauthor{Honghua Zhang}{equal,ucla}
\icmlauthor{Meihua Dang}{equal,ucla}
\icmlauthor{Nanyun Peng}{ucla}
\icmlauthor{Guy Van den Broeck}{ucla}
\end{icmlauthorlist}

\icmlaffiliation{ucla}{Department of Computer Science, University of California, Los Angeles, USA}

\icmlcorrespondingauthor{Honghua Zhang}{hzhang19@cs.ucla.edu}
\icmlcorrespondingauthor{Meihua Dang}{mhdang@cs.ucla.edu}
\icmlcorrespondingauthor{Nanyun Peng}{violetpeng@cs.ucla.edu}
\icmlcorrespondingauthor{Guy Van den Broeck}{guyvdb@cs.ucla.edu}

\icmlkeywords{Machine Learning, ICML}

\vskip 0.3in
]




\printAffiliationsAndNotice{\icmlEqualContribution} 
\begin{abstract}
Despite the success of autoregressive large language models in text generation, it remains a major challenge to generate text that satisfies complex constraints: sampling from the conditional distribution ${\Pr}(\text{text} \given \alpha)$ is intractable for even the simplest lexical constraints $\alpha$. To overcome this challenge, we propose to use tractable probabilistic models~(TPMs) to impose lexical constraints in autoregressive text generation models, which we refer to as~\textbf{\ours}~(\textbf{Ge}nerating \textbf{La}nguage with \textbf{T}ractable C\textbf{o}nstraints). To demonstrate the effectiveness of this framework, we use distilled hidden Markov models, where we \emph{can} efficiently compute ${\Pr}(\text{text} \given \alpha)$, to guide autoregressive generation from GPT2. \ours{} achieves state-of-the-art performance on challenging benchmarks for constrained text generation~(e.g., CommonGen), beating various strong baselines by a large margin. Our work not only opens up new avenues for controlling large language models but also motivates the development of more expressive TPMs.
\end{abstract}

\section{Introduction}
\label{sec:introduction}
Large pre-trained language models (LMs)~\citep{radford2019language, lewis2020bart} have achieved remarkable performance on a wide range of challenging language generation tasks such as machine translation~\citep{bahdanau2015neural, luong2015effective}, summarization~\citep{liu2015toward, xu2019neural} and open-domain creative generation~\citep{yao2019plan,tian2022zero}. Nevertheless, many practical language generation applications require fine-grained control of LMs to follow complex lexical constraints (e.g., given a source document, generate a summary that contains certain keywords). The common paradigm for controlling pre-trained LMs is to either finetune them on task-specific datasets or to condition them on certain prompts. However, finetuning and prompting are by nature approximate solutions and do not guarantee that the desired constraints are satisfied~\citep{meng2022nado, zhang2022paradox}. The major difficulty of constrained language generation lies in the autoregressive nature of LMs: they only model the next token distribution given some prefix $\Pr_{\text{LM}}(x_{t+1} \given x_{1:t})$, while the conditional distribution ${\Pr}_{\text{LM}}(x_{1:n} \given \alpha)$ given a constraint $\alpha$ as simple as, e.g., a keyword appearing at the end of a sentence, is often intractable~\citep{roth1996hardness}.
\begin{figure}[]
    \centering
    \includegraphics[width=0.88\columnwidth]{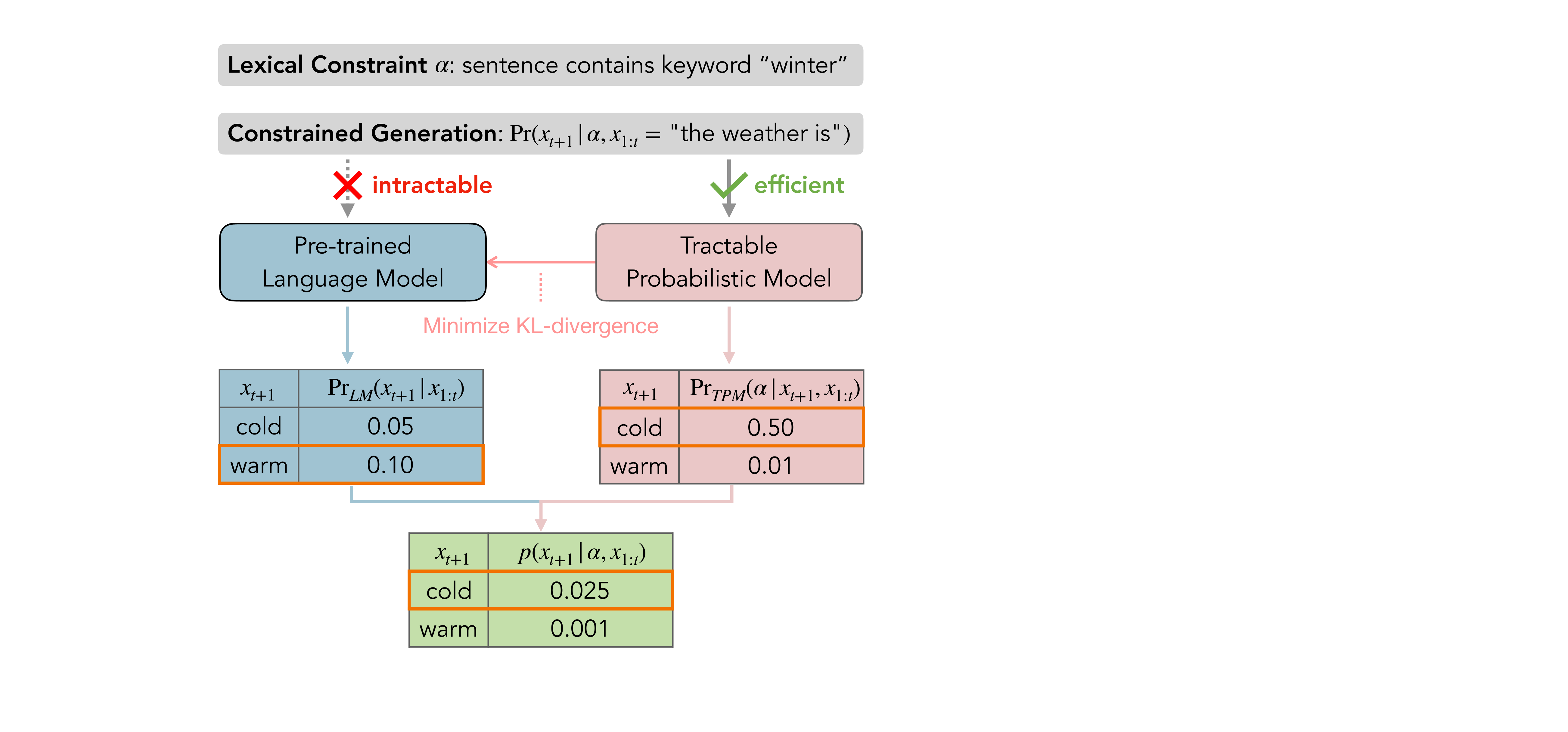}
    \caption{Given some lexical constraint $\alpha$ that we want our pretrained language models to follow in generation, the conditional distribution ${\Pr}(x_{t+1}\given x_{1:t}, \alpha)$ is often intractable. We propose to control and guide the autoregressive generation process of pre-trained LMs via tractable probabilistic models, which do support efficient computation of ${\Pr}(x_{t+1}\given x_{1:t}, \alpha)$.}
    \label{fig:teaser}
\end{figure}

Aside from language models based on neural architectures, one line of research in machine learning focuses on the development of \emph{tractable probabilistic models}~(TPMs)~\citep{poon2011sum,MAL-044,ProbCirc20,ZhangICML21}. TPMs model joint probability distributions and allow for efficient conditioning on various families of logical constraints~\citep{kisa2014probabilistic, ChoiKRR15, BekkerNIPS15}. In this paper, we propose \textbf{\ours{}}~(\textbf{Ge}nerating \textbf{La}nguage with \textbf{T}ractable C\textbf{o}nstraints), where we use TPMs to impose lexical constraints in autoregressive text generation. Given a pre-trained autoregressive LM $\Pr_{\text{LM}}$, e.g., GPT3~\citep{brown2020language}, our goal is to generate text effectively following the conditional distribution $\Pr_{\text{LM}} (x_{1:n} \given \alpha)$ for arbitrary lexical constraints~$\alpha$. As illustrated in Figure~\ref{fig:teaser}, our proposed framework consists of two major components:  (1)~we train a TPM ${\Pr}_{\text{TPM}}$ via maximum likelihood estimation~(MLE) on samples drawn from $\Pr_{\text{LM}}$, which is equivalent to minimizing the KL-divergence between ${\Pr}_{\text{TPM}}$ and $\Pr_{\text{LM}}$; then (2)~ at generation time, we compute ${\Pr}_{\text{TPM}}({x_{t+1} \given x_{1:t}, \alpha})$ efficiently and combine it with $\Pr_{\text{LM}}(x_{t+1} \given x_{1:t})$ to approximate ${\Pr}_{\text{LM}}(x_{t+1} \given x_{1:t}, \alpha)$ for reliable control. Note that we assume nothing about the lexical constraint $\alpha$ as we train $\Pr_{\text{TPM}}$, which means that the TPM does not need to be re-trained for different types of constraints: given a trained TPM that approximates $\Pr_{\text{LM}}$ well enough, we can use it to impose any lexical constraints $\alpha$, as long as $\Pr_{\text{TPM}}( .\given \alpha)$ can be efficiently computed.

Throughout this paper, we use hidden Markov models~(HMMs)~\citep{rabiner1986introduction} as an example TPM to demonstrate the effectiveness of \ours{}. Specifically, (1)~we show that, when trained as probabilistic circuits~\citep{ProbCirc20, liu2022scaling}, HMMs can approximate the GPT2-large model finetuned on downstream tasks well enough and (2)~we propose a dynamic programming algorithm that efficiently computes conditional probabilities ${\Pr}_{\text{HMM}}(\cdot \given \alpha)$, for $\alpha$s that encode constraints as conjunctive normal forms~(CNFs):
$$(I(w_{1,1}) \lor \dots \lor I(w_{1, d_1})) \land \dots \land (I(w_{m,1}) \lor \dots \lor I(w_{m,d_m}));$$
here each $w_{i, j}$ is \emph{a string of tokens}, and $I(w_{i, j})$ is an indicator variable denoting whether or not $w_{ij}$ appears in the generated text. Intuitively, constraint $\alpha$ requires that a set of $m$ keywords must appear somewhere in the generated text, in any of their inflections, where each inflection is encoded as a string of one or more tokens. We evaluate the performance of \ours{} on challenging constrained text generation datasets: CommonGen~\citep{lin2020commongen}, News~\citep{zhang-etal-2020-pointer}, and Yelp!Review~\citep{cho-etal-2019-towards}. \ours{} not only achieves state-of-the-art generation quality but also guarantees that the constraints are satisfied 100\%; for both unsupervised and supervised settings, \ours{} beats strong baselines belonging to different families of constrained generation approaches by a large~margin.

Our study demonstrates the potential of TPMs in controlling large language models and motivates the development of more expressive TPMs.

\section{Guiding Autoregressive Generation with Tractable Probabilistic Models}
\label{sec:auto-generation}
In this section, we present the general \ours{} framework for guiding autoregressive generation with tractable probabilistic models. Throughout this paper, we use uppercase letters $X_t$ for random variables and lowercase letters $x_t$ for their assignment. 

Let ${\Pr}_{\text{LM}}(x_{1:n})$ be the distribution of an autoregressive LM~(e.g., GPT) over $n$ tokens and $\alpha$ a lexical constraint defined over $X_{1:n}$; our goal is to generate from the following conditional distribution: 
$${\Pr}_{\text{LM}}(x_{1:n} \given \alpha) = {\prod}_{t} {\Pr}_{\text{LM}}(x_{t+1} \given x_{1:t}, \alpha)$$ 
Though $\Pr_{\text{LM}}(x_{t+1} \given x_{1:t}, \alpha)$ is intractable, we can assume that $\Pr_{\text{TPM}}(x_{t+1} \given x_{1:t}, \alpha)$ can be efficiently computed.

The first step of \ours{} is to train our TPM model such that $\Pr_{\text{TPM}}$ approximates $\Pr_{\text{LM}}$ as well as possible. We train the TPM model via maximum likelihood estimation~(MLE) on data drawn from $\Pr_{\text{LM}}$, that is, we maximize
\begin{align*}
\mathbb{E}_{x_{1:n} \sim {\Pr}_{\text{LM}}} \log{\Pr}_{\text{TPM}}(x_{1:n}),
\end{align*}
which effectively minimizes their KL-divergence:
\begin{align*}
&D_{\text{KL}}({\Pr}_{\text{LM}}\parallel{\Pr}_{\text{TPM}}) \\
&\!=\!\mathbb{E}_{x_{1:n}\sim {\Pr}_{\text{LM}}} \log{\Pr}_{\text{LM}}(x_{1:n})\!-\!\mathbb{E}_{x_{1:n} \sim {\Pr}_{\text{LM}}} \log{\Pr}_{\text{TPM}}(x_{1:n})
\end{align*}
With the recent development of scaling up TPMs~\citep{chiu2020scaling, DangNeurIPS22, liu2022scaling}, we show in Section~\ref{sec:experiments} that it is possible to train TPMs as good enough approximations of LMs.

Now given some TPM as a good enough approximation for the LM that we want to generate from, we combine both models for constrained generation, where the TPM is responsible for providing guidance on incorporating lexical constraints and LM responsible for generating fluent texts.
To derive our formulation, in addition to lexical constraint $\alpha$, we assume that there exists some ``quality'' constraint $\beta$ such that ${\Pr}_{\text{TPM}}( \given \beta)$ is even closer to ${\Pr}_{\text{LM}}$; intuitively we interpret $\beta$ as some constraint characterizing the high-quality (fluent \& grammatical) sentences that are likely to be sampled from our base LM~$\Pr_{\text{LM}}$. Hence, in order to generate a high-quality sentence satisfying some lexical constraint $\alpha$, we generate from
\begin{align*}
{\Pr}_{\text{TPM}}(x_{1:n} \given \alpha, \beta) = {\prod}_{t} {\Pr}_{\text{TPM}}(x_{t+1} \given x_{1:t}, \alpha, \beta);
\end{align*}
in particular, in addition to the assumption that ${\Pr}_{\text{TPM}}(\cdot \given \beta)$ is a good enough approximation for ${\Pr}_{\text{LM}}$, we also assume the \emph{key independence assumption}: $\alpha$ and $\beta$ are conditionally independent given $x_{1:t+1}$. By applying Bayes rule, it follows from our assumptions that:
\begin{align*}
&{\Pr}_{\text{TPM}}(x_{t+1} \given x_{1:t}, \alpha, \beta) \\
&\quad \propto {\Pr}_{\text{TPM}}(\alpha \given x_{1:t+1}, \beta) \cdot {\Pr}_{\text{TPM}}(x_{t+1} \given x_{1:t}, \beta) \\
&\quad \propto {\Pr}_{\text{TPM}}(\alpha \given x_{1:t+1}) \cdot {\Pr}_{\text{LM}}(x_{t+1} \given x_{1:t}).
\end{align*}

Now we examine whether our key independence assumption holds for the \emph{unsupervised} and \emph{supervised} settings.

\textbf{Unsupervised setting.} In the unsupervised setting, we assume that the base pre-trained LM is \emph{not} finetuned given task-specific supervision; that is, $\Pr_{\text{LM}}$ is not finetuned to generate texts satisfying $\alpha$ provided as input, but is possibly finetuned or prompted for the purpose of domain adaptation. In this setting, there is no easy way for the ``quality'' constraint $\beta$ to obtain any information about the lexical constraint $\alpha$ and our key independence assumption should roughly hold. In other words, satisfying the lexical constraint $\alpha$ should not help or hinder the fluency of the generated sentence according to the pre-trained LM, it merely biases what the sentence talks about.
Hence for the unsupervised setting, we generate autoregressively following the next-token distribution defined as:
\begin{equation}    
p(x_{t+1} \!\given x_{1:t}, \alpha) \propto {\Pr}_{\text{TPM}}(\alpha \given x_{1:t+1}) \cdot {\Pr}_{\text{LM}}(x_{t+1} \given x_{1:t}).
\label{eq:unsupervised}
\end{equation}
This formulation is also adopted in FUDGE~\citep{yang2021fudge} and NADO~\citep{meng2022nado}, which train auxiliary models to approximate $\Pr_{\text{LM}}(\alpha \given x_{1:t+1})$; the key difference is that such auxiliary models take $\alpha$ as input during training while our TPM training is unconditional.

\textbf{Supervised setting.} In this setting, we assume that the language model ${\Pr}_{\text{LM}}$ is finetuned in a sequence-to-sequence~(seq2seq) manner; that is, during training, $\alpha$ is explicitly supplied to the LM together with some gold sentences: e.g., for keyword-type constraints, the LM is finetuned over texts of the form \emph{``weather winter cold = the weather is cold in winter,''} where the prompt \emph{``weather winter cold = ''} encodes the constraint that all words before \emph{``=''} should be used. In this case, our key independence assumption no longer holds because ${\Pr}_{\text{LM}}$ is already trained to satisfy the lexical constraint $\alpha$, which is provided as part of the prefix $x_{1:t+1}$. 
Hence for the supervised setting, we adopt an alternative formulation by viewing ${\Pr}_{\text{TPM}}(x_{t+1} \given x_{1:t}, \alpha)$ and ${\Pr}_{\text{LM}}(x_{t+1} \given x_{1:t})$ as classifiers trained for the same task yet with different biases; by \citet{satopaa2014combining}, if we assume that each model predicts the true logits up to additive Gaussian noise, then the most likely logits can be found by taking a geometric mean of the models. Hence, in the supervised setting, we generate autoregressively following the next-token distribution defined as their weighted geometric mean~\citep{hinton2002training, grover2018boosted}:
\begin{align}
&p(x_{t+1} \given x_{1:t}, \alpha) \nonumber \\
&\quad \propto {\Pr}_{\text{TPM}}(x_{t+1} \given x_{1:t}, \alpha)^{w}\!\cdot\!{\Pr}_{\text{LM}}(x_{t+1}\given x_{1:t})^{1-w};
\label{eq:supervised}
\end{align}
here $w \in (0, 1)$ is a hyper-parameter to be tuned.

To summarize, \ours{} consists of two major steps: (1)~distillation: we train a TPM on samples drawn from the pre-trained LM via MLE to effectively minimize the KL divergence between $\Pr_{\text{LM}}$ and $\Pr_{\text{TPM}}$; (2)~probabilistic reasoning: for each step of autoregressive generation, we compute $\Pr_{\text{TPM}}(\cdot \given \alpha)$ and generate from the conditional next-token distribution~$p(x_{t+1} \given x_{1:t}, \alpha)$ defined above. In addition to better generation quality, which we demonstrate in Section~\ref{sec:experiments}, \ours{} has two major advantages compared to its counterparts for constrained generation:
\begin{itemize}
    \item The sentences generated following $p(x_{t+1} \given x_{1:t}, \alpha)$ are \emph{guaranteed} to satisfy the lexical constraint $\alpha$; in autoregressive generation, as we generate the next token $x_{t+1}$, it follows from the definition that for choices of $x_{t+1}$ such that \emph{$\alpha$ cannot be satisfied}, ${\Pr}_{\text{TPM}}(x_{t+1}, x_{1:t}, \alpha)$ is $0$, thus $p(x_{t+1} \given x_{1:t}, \alpha)$ is also $0$.
    \item The TPM training is independent of the lexical constraint $\alpha$, which is only enforced at inference time; it immediately follows that we do not need to re-train the TPM model no matter how $\alpha$ changes; on the other hand, constrained decoding approaches that train auxiliary neural models, e.g., FUDGE and NADO, need to re-train their model for different types of constraints.
\end{itemize}

Throughout the rest of this paper, we use \textbf{hidden Markov models}~(HMMs) as example TPMs to demonstrate the practicality and effectiveness of \ours{}. In the following section, we propose an efficient algorithm for computing ${\Pr}_{\text{TPM}}(\alpha \given x_{1:t+1})$ and ${\Pr}_{\text{TPM}}(x_{t+1} \given x_{1:t}, \alpha)$.

\section{Efficient Probabilistic Reasoning with Hidden Markov Models}
\label{sec:dp}
\begin{figure*}
    \centering
    \includegraphics[width=0.92\linewidth]{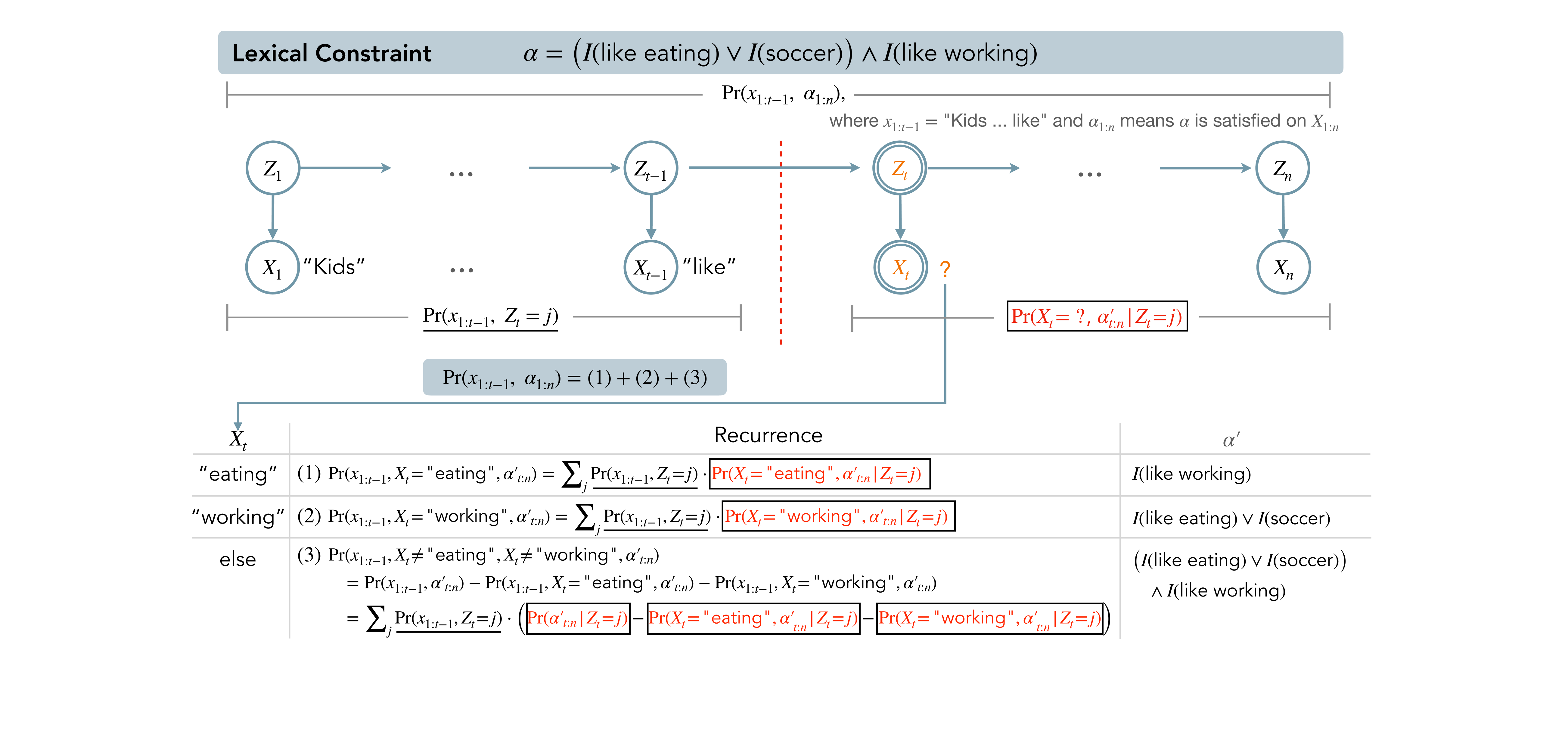}
    \caption{A toy example illustrating our dynamic programming algorithm. Here, given the the first $t\!-\!1$ tokens ``Kids ... like'' that have been generated, the figure illustrates how to compute $\Pr(X_{1:{t-1}}\!=\!\text{``Kids ... like''}, \alpha_{1:n})$. We consider three possible cases for the next token $X_t$:~``eating'', ``working'' or neither, and for each case we can reduce the constraint $\alpha_{1:n}$ to the ``easier'' constraint ${{\alpha}^{\prime}}_{t:n}$ for some $\alpha^{\prime}$. Then by conditioning on $Z_t = j$ (hidden states), we can break down $\Pr(x_{1:t-1}, X_{t} = ?, {\alpha^{\prime}}_{t:n})$ into two terms: $\Pr(x_{1:t-1}, Z_{t} = j)$ and $\Pr(X_{t} = ?, {\alpha^{\prime}}_{t:n} \given Z_{t}=j)$, which are underlined and boxed in the figure, respectively; in particular the underlined terms can be computed by the forward algorithm for HMMs and the boxed terms can be computed recursively by the dynamic programming algorithm.
    }
    \label{fig:case}
\end{figure*}
To impose lexical constraint $\alpha$ in autoregressive generation via TPM, for any given prefix $x_{1:t}$, we need to compute ${\Pr}_{\text{TPM}}(x_{1:t}, \alpha)$~(omit subscript for rest of the section); specifically, as described in Section~\ref{sec:auto-generation}, we need to compute $\Pr_{}(\alpha \given x_{1:t+1})\!=\!\Pr_{}(x_{1:t+1}, \alpha) / \Pr_{}(x_{1:t+1})$ for the unsupervised setting and $\Pr_{}(x_{t+1} \given x_{1:t}, \alpha) \propto \Pr(x_{1:t+1}, \alpha)$ for the supervised setting.
In this section, we describe a dynamic programming algorithm that computes $\Pr(x_{1:t}, \alpha)$ for hidden Markov models~(HMMs), where $\alpha$ is some lexical constraint encoded in a conjunctive normal form~(CNF):
$$(I(w_{1,1}) \lor \dots \lor I(w_{1, d_1})) \land \dots \land (I(w_{m,1}) \lor \dots \lor I(w_{m,d_m}));$$
here each $w_{i,j}$ is a string of tokens, which we denote as ``keystrings'' for short, and $I(w_{ij})$ is the indicator variable that represents whether $w_{ij}$ appears in the generated text.\footnote{To be precise, denoting the $k$th token of $w_{ij}$ as $(w_{ij})_{k}$, $I(w_{i,j})$ is in fact a disjunction over conjunctions: ${\lor}_{1\leq t\leq n-|w_{ij}|+1} \left({\land}_{0 \leq k < |w_{ij}|} X_{t+k}=(w_{ij})_{k}\right)$, representing that $w_{ij}$ can be in any position of the generated text.} We refer to $(I(w_{i,1}) \lor \dots \lor I(w_{i, d_{i}}))$ as a \emph{clause}.

For simplicity, we use the short-hand $\alpha_{l:r}$ to denote the event that $\alpha$ is satisfied on the \emph{sub-sequence $X_{l:r}$}. In practice, we treat HMMs as language models over sequences of tokens of maximum length $n$ and the lexical constraint we enforce is denoted as $\alpha_{1:n}$; in the following discussions, we write $\Pr(x_{1:t}, \alpha_{1:n})$ instead of $\Pr(x_{1:t}, \alpha)$.

\subsection{Hidden Markov Models}
A hidden Markov model~(HMM) represents a joint probability distribution over $n$ observed variables $X_{1:n}$ and $n$ latent variables $Z_{1:n}$. Specifically, for language modeling, $X_{t}$ represents the token at position $t$ and $Z_{t}$ represents the corresponding latent state; $Z_{t}$ takes values in $\{1, 2, \dots, h\}$, where $h$ is the \emph{number of latent states}. Given observed token sequence $x_{1:n}$ and latent state sequence $z_{1:n}$, the joint probability $\Pr(x_{1:n}, z_{1:n})$ is defined as: 
\begin{align*}
\Pr(x_{1} \given z_{1})\Pr(z_{1}) {\prod}_{2 \leq t \leq n} \Pr(x_{t} \given z_{t}) \Pr(z_{t} \given z_{t-1});
\end{align*}
in particular, the parameters of HMM are given by the initial probability $\Pr(z_{1})$, emission matrix $\Pr(x_{t} \given z_{t})$ and the transition matrix $\Pr(z_{t+1} \given z_{t})$, which stay the same across different positions $t$. HMMs can also be represented as Bayesian networks~\citep{pearl1985bayesian}; see Figure~\ref{fig:case} for an example. To perform probabilistic inference on HMMs efficiently, we leverage the following \emph{Markov property}:
\begin{align}
\label{eq:independence}
\Pr(x_{t:n} \given z_{t}, x_{1:t-1}) = \Pr(x_{t:n} \given z_{t}).
\end{align}
For example, we can compute the probability of any prefix $\Pr(x_{1:t}) = \sum_{z_t} \Pr(x_{1:t}, z_t)$, which can be efficiently computed by the following recurrence relation, which is referred to as the \emph{forward algorithm}~\citep{rabiner1986introduction}:
\begin{align*}
\begin{split}
\label{eq:forward}
&\Pr(x_{1:t}, z_{t}) \\
&\text{ }\!=\!\sum_{1\leq z_{t-1}\leq h}\Pr(x_{t} \given z_{t})\Pr(z_{t} \given z_{t-1})\Pr(x_{1:t-1}, z_{t-1}).
\end{split}
\end{align*}

\textbf{Modeling Variable-length Texts with HMMs.} HMMs model distributions over a fixed number of random variables $X_{1:n}$. To model texts with variable lengths, we first determine a maximum sequence length $n$ and pad training texts of length $< n$ with the special EOS (``endoftext'') token to the maximum length. We also construct our HMM in a special way such that an EOS token can only be followed by EOS tokens; that is, sequences that do not satisify this constraint have $0$ probability. Hence, ${\Pr}_{\text{HMM}}(x_{1:n})$ effectively defines a distribution over all texts with length $\leq n$.

\subsection{An Efficient Dynamic Programming Algorithm}
We first illustrate the dynamic programming algorithm with a toy example. As shown in Figure~\ref{fig:case}, assume that we have generated the first $t-1$ tokens ``Kids ... like'' and we are given the constraint:
$$\alpha = I(\text{like} \oplus \text{working}) \land \left(I(\text{like} \oplus \text{eating}) \lor I(\text{soccer})\right);$$
here we assume ``like'', ``working'', ``eating'' and ``soccer'' are single tokens and $\oplus$ denotes string concatenation. To compute $\Pr(x_{1:t-1}, \alpha_{1:n})$, we marginalize out $x_t$ in $\Pr(x_{1:t-1}, x_{t}, \alpha_{1:n})$; in particular, we sum over three possible cases for the next token $x_t$: ``eating'', ``working'' or neither, and for each case we reduce $\Pr(x_{1:t-1}, x_{t}, \alpha_{1:n})$ to $\Pr(x_{1:t-1}, x_{t}, \alpha^{\prime}_{t:n})$ for some $\alpha^{\prime}$; here $\alpha^{\prime}$ is some CNF formula obtained by removing from the original $\alpha$ the clauses that are already satisfied. Then, we leverage the Markov property of HMMs~(see Equation~\ref{eq:independence}) to break down the joint probability $\Pr(x_{1:t-1}, x_{t}, \alpha^{\prime}_{t:n})$ into sub-problems. 

\begin{algorithm}[tb]
   \caption{Constrained Sampling with GeLaTo}
   \label{alg:dp}
\begin{algorithmic}
   \STATE {\bfseries Input:} constraint~$\alpha$, maximum text length~$n$, HMM~$q_1$, autoregressive LM~$q_2$, \# of HMM latent states~$h$.
   \FOR{$l$ {\bfseries from} $n$ {\bfseries to} $1$} 
       \FOR{$x$ {\bfseries in} suffixes of keystrings in $\alpha$, $z_l$ {\bfseries from} $1$ to {\bfseries $h$}}
           \FOR{$\psi$ {\bfseries in} subsets of clauses of $\alpha$}
               \STATE compute $q_1(x_{l:r}, \psi_{l:n} \given z_l)$ by the recurrence relation and store values in cache.
           \ENDFOR
       \ENDFOR
   \ENDFOR
   \STATE {\bfseries initialize} $x_{1:0} = \text{empty string}$
   \FOR{$t$ {\bfseries from} $1$ {\bfseries to} $n$} 
       \FOR{$x_{t}$ {\bfseries in} \text{vocabulary}}
           \STATE compute $q_1(\alpha \given x_{1:t-1}, x_t)$ by Case 1.
           \STATE $p(x_{t} \given x_{1:t-1}, \alpha)\!=\!{q_1}(\alpha \given x_{1:t-1}, x_t) q_{2}(x_t \given x_{1:t-1})$
       \ENDFOR
       \STATE sample $x_{t} \sim p(\cdot \given x_{1:t-1}, \alpha)$
       \STATE update $x_{1:t} := x_{1:t-1} \oplus x_{t}$
   \ENDFOR
   \STATE {\bfseries return} $x_{1:n}$
\end{algorithmic}
\end{algorithm}

Before we describe how to compute $\Pr(x_{1:t}, \alpha_{1:n})$, we establish a recurrence relation for computing terms of the form $\Pr(x_{l:r},\psi_{l:n} \given z_l)$, where $x_{l:r}$ is either the empty string or a suffix for some keystring in $\alpha$, $\psi$ is a CNF consisting of a subset of clauses in $\alpha$ and $z_{l}$ is a latent state for $Z_{l}$.

\textbf{Assumptions \& Notations.} For simplicity, we make the following \emph{non-overlapping assumption}: for the set of keystrings appearing in $\alpha$, denoted as $\{w_{ij}\}$, the prefix of $w_{ij}$ cannot be a suffix for $w_{pq}$ for all $ij \neq pq$. We also define the following set of strings:
$$S(x, \alpha) := \{s : \exists x^{\prime} \text{ a suffix of } x \text{ s.t. } x^{\prime} \oplus s \text{ lies in } \alpha \},$$
which contains all strings that can be appended to $x$ to form some keystrings in $\alpha$. For the example in Figure~\ref{fig:case}, for $x = \text{``Kids ... like''}$, $S(x, \alpha)$ is given by $\{\text{``eating''}, \text{``working''}\}$. We write $s_{i:j}$ as a shorthand for $X_{i:j} = s$.

\textbf{Recurrence Relation.} $\Pr(x_{l:r},\psi_{l:n} \given z_l)$ follows the following recurrence relation:

Case 1. $x_{l:r} \neq \emptyset$; then,
\begin{align*}
&\Pr(x_{l:r}, \alpha_{l:n} \given z_l) \\
&\text{ } = \sum_{z_{r+1}} \underline{\Pr(x_{l:r},z_{r+1} \given z_l)} \bigg(\boxed{\Pr(\alpha_{r+1:n} \given z_{r+1})}  \\
&\text{ }\text{ } + \sum_{s \in S(x_{l:r}, \alpha)} \boxed{\Pr(s_{r+1:r+|s|}, (\alpha \setminus x_{l:r}\oplus s)_{r+1:n} \given z_{r+1})} \\
&\text{ }\text{ } - \sum_{s \in S(x_{l:r}, \alpha)} \boxed{\Pr(s_{r+1:r+|s|}, \alpha_{r+1:n} \given z_{r+1})}\bigg);
\end{align*}
here $\oplus$ denotes string concatenation and $\alpha\setminus x_{l:r}\oplus s$ represents the CNF obtained by removing the clauses with any keywords appearing in $x_{l:r}\oplus s$.

Case 2. $x_{l:r} = \emptyset$; we reduce the problem to Case 1 by enumerating $x_l$ over the vocabulary:
\begin{align*}
&\Pr(\alpha_{l:n} \given z_{l}) = \sum_{x_l \in \text{vocabulary}} \boxed{\Pr(x_l, \alpha_{l:n} \given z_{l})};
\end{align*}

The recurrence relation presented above gives us a dynamic programming algorithm for computing terms of the form $\Pr(x_{l:r},\psi_{l:n} \given z_l)$; see appendix for derivations. Note that the boxed terms are the sub-problems and the underlined terms are either HMM parameters or can be pre-computed via the forward algorithm and then cached for later use. 

Finally, as discussed at the beginning of this section, we guide autoregressive generation from language models at step $t$ by computing $\Pr(x_{1:t-1}, x_{t}, \alpha_{1:n})$, where $x_{1:t-1}$ denotes the first $t-1$ tokens that have been generated:
\begin{align*}
\Pr(x_{1:t}, \alpha_{1:n}) = {\sum}_{z_1} \Pr(z_1) \Pr(x_{1:t}, \alpha_{1:n} \given z_1);
\end{align*}
here $\Pr(z_1)$ is the initial probability of the HMM and $\Pr(x_{1:t}, \alpha_{1:n} \given z_1)$ can be computed by the formula in Case~1~(setting $l\!=\!1$), given that all boxed terms are pre-computed by the dynamic programming algorithm.

As an example, Algorithm~\ref{alg:dp} summarizes how to perform constrained generation with GeLaTo by sampling from the autoregressive distribution $p(x_{t} \given x_{1:t-1}, \alpha)$, as defined in Section~\ref{sec:auto-generation}~(unsupervised setting). We can easily adapt Algorithm~\ref{alg:dp} for other decoding procedures like beam search. 

For a rough analysis of the time complexity of Algorithm~\ref{alg:dp}, we treat both the number of latent states $h$ and the vocabulary size as constants; in practice, we can avoid enumerating all tokens in the vocabulary and all latent states of HMM via GPU parallelization~(see appendix \& code for details). It follows that the time complexity of GeLaTo is $O(2^{|\alpha|}nm)$, where $|\alpha|$ is the number of clauses in $\alpha$, $n$ is the maximum sequence length and $m$ is the number of different suffixes for all keystrings in $\alpha$. 
We show that \ours{} scales well in practice in Section~\ref{subsec:results}.

\section{Experiments}
\label{sec:experiments}
In this section, we demonstrate the effectiveness of \ours{}\footnote{\url{https://github.com/UCLA-StarAI/GeLaTo}} on challenging benchmarks for constrained generation: CommonGen~\citep{lin2020commongen}, Yelp!Review~\citep{cho-etal-2019-towards} and News~\citep{zhang-etal-2020-pointer}; in particular, we focus on CommonGen for detailed analysis. For both unsupervised and supervised settings, \ours{} achieves state-of-the-art performance in terms of various automatic evaluation metrics including BLEU score while guaranteeing 100\% constraint satisfaction. 

\begin{table*}[!th]
    \centering
    \footnotesize
\begin{tabular}{l ||cc|cc|cc|cc|cc|cc}
\toprule
\multicolumn{1}{c||}{\multirow{2}{*}{Method}} & \multicolumn{8}{c|}{{\it Generation Quality}}        & \multicolumn{4}{c}{{\it Constraint Satisfaction}} \\
\multicolumn{1}{c||}{}              & \multicolumn{2}{c}{ROUGE-L} & \multicolumn{2}{c}{BLEU-4} & \multicolumn{2}{c}{CIDEr} & \multicolumn{2}{c|}{SPICE} & \multicolumn{2}{c}{Coverage}            & \multicolumn{2}{c}{Success Rate}           \\\hline

\multicolumn{1}{l}{\cellcolor{gray!25} \textit{Unsupervised}}
& \multicolumn{1}{c}{\cellcolor{gray!25} \textit{dev}} &\multicolumn{1}{c}{\cellcolor{gray!25} \textit{test}}
& \multicolumn{1}{c}{\cellcolor{gray!25} \textit{dev}} &\multicolumn{1}{c}{\cellcolor{gray!25} \textit{test}}
& \multicolumn{1}{c}{\cellcolor{gray!25} \textit{dev}} &\multicolumn{1}{c}{\cellcolor{gray!25} \textit{test}}
& \multicolumn{1}{c}{\cellcolor{gray!25} \textit{dev}} &\multicolumn{1}{c}{\cellcolor{gray!25} \textit{test}}
& \multicolumn{1}{c}{\cellcolor{gray!25} \textit{dev}} &\multicolumn{1}{c}{\cellcolor{gray!25} \textit{test}}
& \multicolumn{1}{c}{\cellcolor{gray!25} \textit{dev}} &\multicolumn{1}{c}{\cellcolor{gray!25} \textit{test}}\\

InsNet~\citep{lu2022insnet}         &-&-      &18.7&-       &-&-        &-&-    &\textbf{100.0}&-   &\textbf{100.0}&-\\ 
NeuroLogic~\citep{lu2021neurologic} &-&41.9   &-&24.7       &-&14.4     &-&27.5    &-&~96.7       &-&-          \\ 
A*esque~\citep{lu2022astar}         &-&\textbf{44.3}   &-&28.6  &-&\textbf{15.6}    &-&29.6       &-&~97.1       &-&-          \\
NADO~\cite{meng2022nado}            &-&-      &26.2&-       &-&-    &-&-    &~96.1&-       &-&-              \\
\ours                           &\textbf{44.3}&43.8   &\textbf{30.3}&\textbf{29.0} &\textbf{15.6}&15.5    &\textbf{30.2}&\textbf{30.3}        &\textbf{100.0}&\textbf{100.0}      &\textbf{100.0}&\textbf{100.0}\\

\multicolumn{1}{l}{\cellcolor{gray!25} \textit{Supervised}}& \multicolumn{1}{c}{\cellcolor{gray!25} \textit{dev}} &\multicolumn{1}{c}{\cellcolor{gray!25} \textit{test}}& \multicolumn{1}{c}{\cellcolor{gray!25} \textit{dev}} &\multicolumn{1}{c}{\cellcolor{gray!25} \textit{test}}
& \multicolumn{1}{c}{\cellcolor{gray!25} \textit{dev}} &\multicolumn{1}{c}{\cellcolor{gray!25} \textit{test}}
& \multicolumn{1}{c}{\cellcolor{gray!25} \textit{dev}} &\multicolumn{1}{c}{\cellcolor{gray!25} \textit{test}}
& \multicolumn{1}{c}{\cellcolor{gray!25} \textit{dev}} &\multicolumn{1}{c}{\cellcolor{gray!25} \textit{test}}
& \multicolumn{1}{c}{\cellcolor{gray!25} \textit{dev}} &\multicolumn{1}{c}{\cellcolor{gray!25} \textit{test}}\\
NeuroLogic~\citep{lu2021neurologic} &-&42.8   &-&26.7       &-&14.7        &-&30.5    &-&97.7       &-&93.9$^\dag$          \\ 
A*esque~\citep{lu2022astar}         &-&43.6   &-&28.2       &-&15.2        &-&30.8    &-&97.8       &-&97.9$^\dag$           \\ 
NADO~\cite{meng2022nado}            &44.4$^\dag$&-   &30.8&-       &16.1$^\dag$&-        &32.0$^\dag$&-    &97.1&-       &88.8$^\dag$ &-          \\ 
\ours                           &\textbf{46.2}&\textbf{45.9}  &\textbf{34.0}&\textbf{34.1} &\textbf{17.2}&\textbf{17.5}        &\textbf{32.2}&\textbf{33.5}              &\textbf{100.0}&\textbf{100.0}     &\textbf{100.0} &\textbf{100.0}\\
\bottomrule
\end{tabular}
    \caption{
    Performance comparison of different generation methods for \emph{unsupervised} and \emph{supervised} settings on the CommonGen dataset, measured by \emph{generation quality} and \emph{constraint satisfaction}. For hyper-parameter tuning, we conduct cross-validation on a small subset of the training set and report evaluation results for both validation (\textit{dev}) and {\textit test} set. All methods except for InsNet uses GPT2-large as their base model.
    Numbers with $\dag$ are reproduced by ourselves.
    }
    \label{tab:sota}
\end{table*}


\subsection{Dataset \& Baselines}
\label{subsec:dataset}
CommonGen~\citep{lin2020commongen} is a benchmark for constrained generation with lexical constraints: the input of each example consists of three to five concepts~(keywords) and the goal is to generate a natural sentence using all concepts; in particular, the given keywords can appear in any order or in any form of inflections in the generated sentences. For example, given \emph{``car snow drive''} as concepts, both \emph{``a man drives a car on a snow covered road''} and \emph{``the car drove through the snow''} are considered acceptable. We also evaluate \ours{} on the Yelp!Review~\citep{cho-etal-2019-towards} and the News~\citep{zhang-etal-2020-pointer} datasets. Compared to CommonGen, both Yelp!Review and News share similar formats, except that they require all keywords to be generated in the forms as given~(i.e.\ no inflections allowed) and to follow specific orders.

We compare \ours{} against constrained generation approaches belonging to different families: 

\textbf{InsNet}~\citep{lu2022insnet} is a class of insertion-based language models~\citep{susanto2020lexically} that generate text by repeatedly inserting new tokens into the sequence. InsNet guarantees that the keywords appear in the generated sentence by initializing the token sequence as the keywords, arranged in some order.

\textbf{NeuroLogic (A*esque) Decoding}~\citep{lu2021neurologic, lu2022astar} are search-based decoding algorithms; they are inference-time algorithms like beam search and do not use any auxiliary models. Leveraging look-ahead heuristics, NeuroLogic A*esque decoding not only optimizes the probability of the generated sentence but also steers the generation towards satisfying the lexical constraints. 

\textbf{NADO}~\citep{meng2022nado} trains an auxiliary neural model approximating the conditional distribution $\Pr(\alpha | x_{1:t}, x_{t+1})$ to guide constrained generation of the base model. As mentioned in Section~\ref{sec:auto-generation}, NADO needs to re-train the auxiliary model for different types of $\alpha$ (e.g., ten keywords) while \ours{} does not need re-training.

\subsection{Approach}
Following the experiment setup of \citet{lu2021neurologic} and \citet{meng2022nado}, we evaluate \ours{} under both unsupervised and supervised settings, as described in Section~\ref{sec:auto-generation}. 

\paragraph{finetuning GPT2-large} All baselines, except for InsNet, perform generation with GPT2-large~\citep{radford2019language} as the \emph{base model}. Following prior works~\citep{meng2022nado}, we use finetuned GPT2-large as base models:
\begin{enumerate}
    \item Unsupervised Setting: we perform \emph{domain adaptation}~(DA) by finetuning GPT2-large on all gold~(reference) sentences of the training split of CommonGen \emph{without} supplementing the keywords. We finetune the model for 1 epoch with learning rate = 1e-6.
    \item Supervised Setting: following the template proposed in~\citet{lin2020commongen}, we finetune the GPT2-large model in a \emph{sequence-to-sequence}~(seq2seq) manner; in particular we finetune the model on sequences of the form \emph{``car snow drive = a car drove through snow''} for 3 epochs with learning rate = 1e-6.
\end{enumerate}

\textbf{Training HMMs.} We use HMMs as an example TPM to enforce lexical constraint in autoregressive generation from GPT2-large. Following Section~\ref{sec:experiments}, we sample sequences of length 32 from the finetuned GPT2-large models and train HMMs with 4096 hidden states to approximate the base model distributions; we train HMMs with the expectation-maximization~(EM) algorithm for 40 epochs, and we re-sample 0.2 million examples for each epoch. The HMM models are trained as probabilistic circuits with the \emph{Juice.jl} framework~\citep{DangAAAI21} and the training procedure leverages the latent variable distillation technique proposed in~\citet{liu2022scaling}; we refer readers to the original papers for more details.

\textbf{Constraint Formulation.} For CommonGen, as described in Section~\ref{subsec:dataset}, the goal is to generate a sentence using the given concepts~(keywords) and we encode this lexical constraint as a CNF. For example, given the concepts ``catch frisbee snow'', the lexical constraint can be represented as:
\begin{align*}
    & [ I(\text{catch}) \lor I(\text{caught}) \lor \dots ] \\
    \land &[ I(\text{fr} \oplus \text{is} \oplus \text{bee}) \lor I(\text{fr} \oplus \text{is} \oplus \text{bees}) \lor \dots ] \\
    \land &[ I(\text{snow}) \lor I(\text{snow} \oplus \text{ing}) \lor I(\text{snow} \oplus \text{ed}) \lor \dots ];
\end{align*}
here each clause encodes the constraint that a keyword has to appear, in any form of its inflections; each literal $I(w)$ indicates the occurrence of a string of tokens $w$~(i.e.\ keystring), which represents the tokenization of a specific inflection of a keyword and $\oplus$ denotes the concatenation of individual tokens. For the keywords, we use LemmInflect\footnote{\url{https://github.com/bjascob/LemmInflect}} to generate their inflections. We also enforce the constraint that each keystring, whenever it appears in the generated text, is followed by either a space, a comma or an $\langle\text{eos}\rangle$ token.

\textbf{Decoding.} $p(x_{t+1} \given x_{1:t}, \alpha)$ defined in Section~\ref{sec:auto-generation}~(see Eq.~\ref{eq:unsupervised} and~\ref{eq:supervised}) induces the conditional distribution $p(x_{1:n} \given \alpha) = \prod_{t} p(x_{t+1} \given x_{1:t}, \alpha)$. We adopt beam search to greedily search for $x_{1:n}$ that maximizes $p(x_{1:n} \given \alpha)$; we experiment with different beam sizes: 16, 32, 64 and 128. Finally, we re-rank all beams generated by beam search by their log-likelihood given by the domain-adapted GPT2-large model and select the top beam.

\textbf{Metrics.} We evaluate the quality of generation via human evaluation and some commonly used automatic metrics including ROUGE~\citep{lin2003automatic}, BLEU~\citep{papineni2002bleu}, CIDEr~\citep{vedantam2015cider}, and SPICE~\citep{anderson2016spice}. In addition to generation quality, we also measure the constraint satisfaction performance via \emph{coverage}, the average percentage of concepts presented in the generated sentences and \emph{success rate}, the percentage of generated sentences that perfectly satisfy the constraints.

\subsection{Results and Analysis}
\label{subsec:results}
Main evaluation results are presented in Table~\ref{tab:sota}. \ours{} outperforms all baselines in both unsupervised and supervised settings by a large margin, achieving not only significantly higher BLEU and ROUGE scores but also 100\% constraint satisfaction. 
The unsupervised setting is more challenging given that the base model is never trained with task-specific supervision; despite this, \ours{} achieves 30.3 BLEU score in the \emph{unsupervised} setting, while NADO~(the best performing baseline) obtains 30.8 BLEU score in the \emph{supervised} setting. To provide more insight into \ours{}, we also conduct the following ablation studies.

\begin{figure}[th]
    \centering
    \includegraphics[width=0.75\columnwidth]{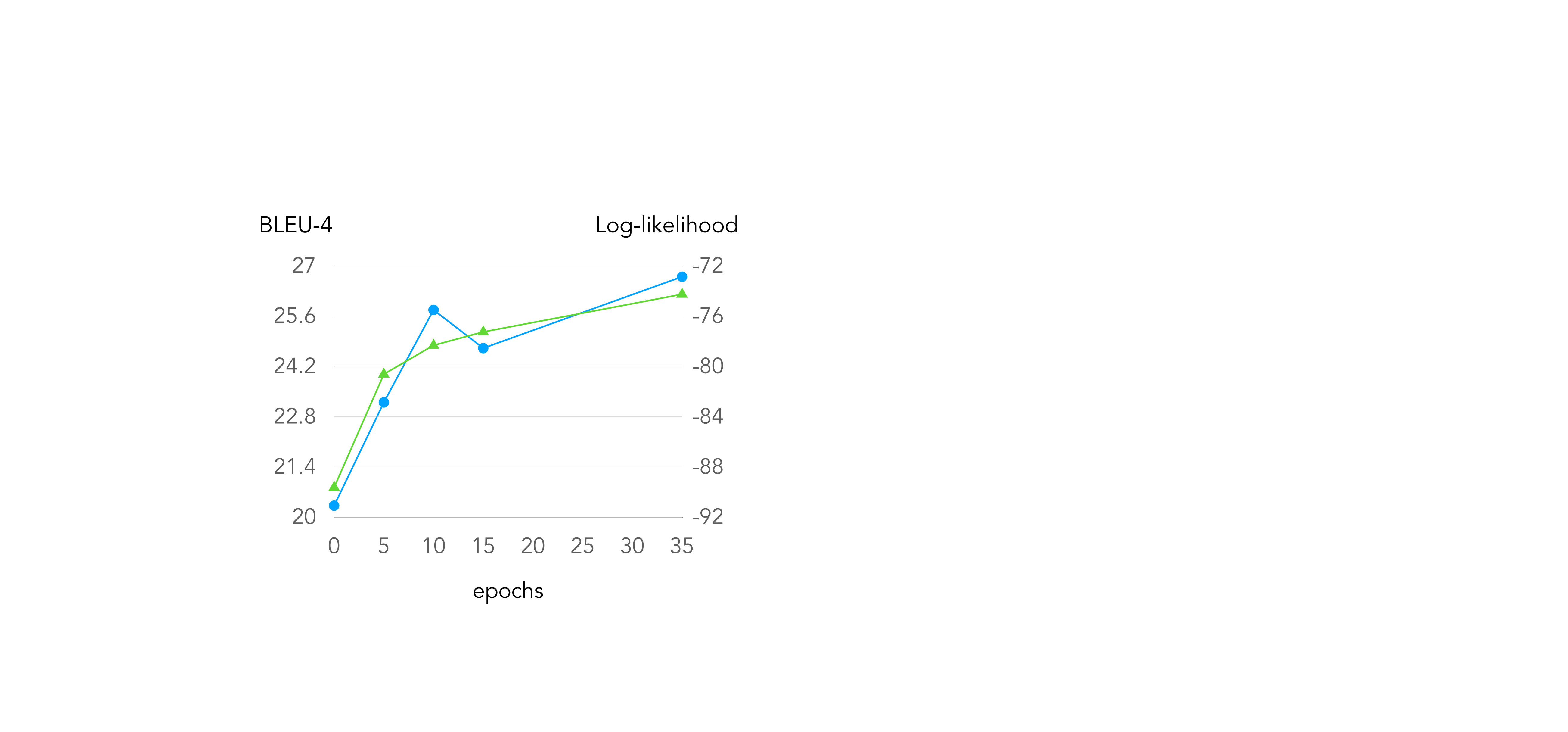}
    \caption{\label{fig:ll-vs-bleu} 
     HMM log-likelihoods on data sampled from GPT-2 large~(triangles) and the corresponding BLEU scores~(circles) w.r.t. \# of training epochs. As the HMM model approximates GPT2-large better, the generation quality also improves. 
    }
\end{figure}

\textbf{Generation Quality vs.\ Approximation Performance.} As discussed in Section~\ref{sec:auto-generation}, \ours{} assumes that distilled HMMs are good enough approximations for base models; our hypothesis is that the better the HMM approximates the base model, the better the generation quality. With \ours{}, we generate from different HMM checkpoints from the distillation procedure, and report the average log-likelihoods and BLEU scores~(without re-ranking the beams). As shown in Figure~\ref{fig:ll-vs-bleu}, as the training proceeds, both log-likelihood and BLEU score improves, exhibiting a clear positive correlation. This finding motivates the development of better tractable probabilistic models for language modeling.

\begin{figure}[t]
    \centering
    \includegraphics[width=0.75\columnwidth]{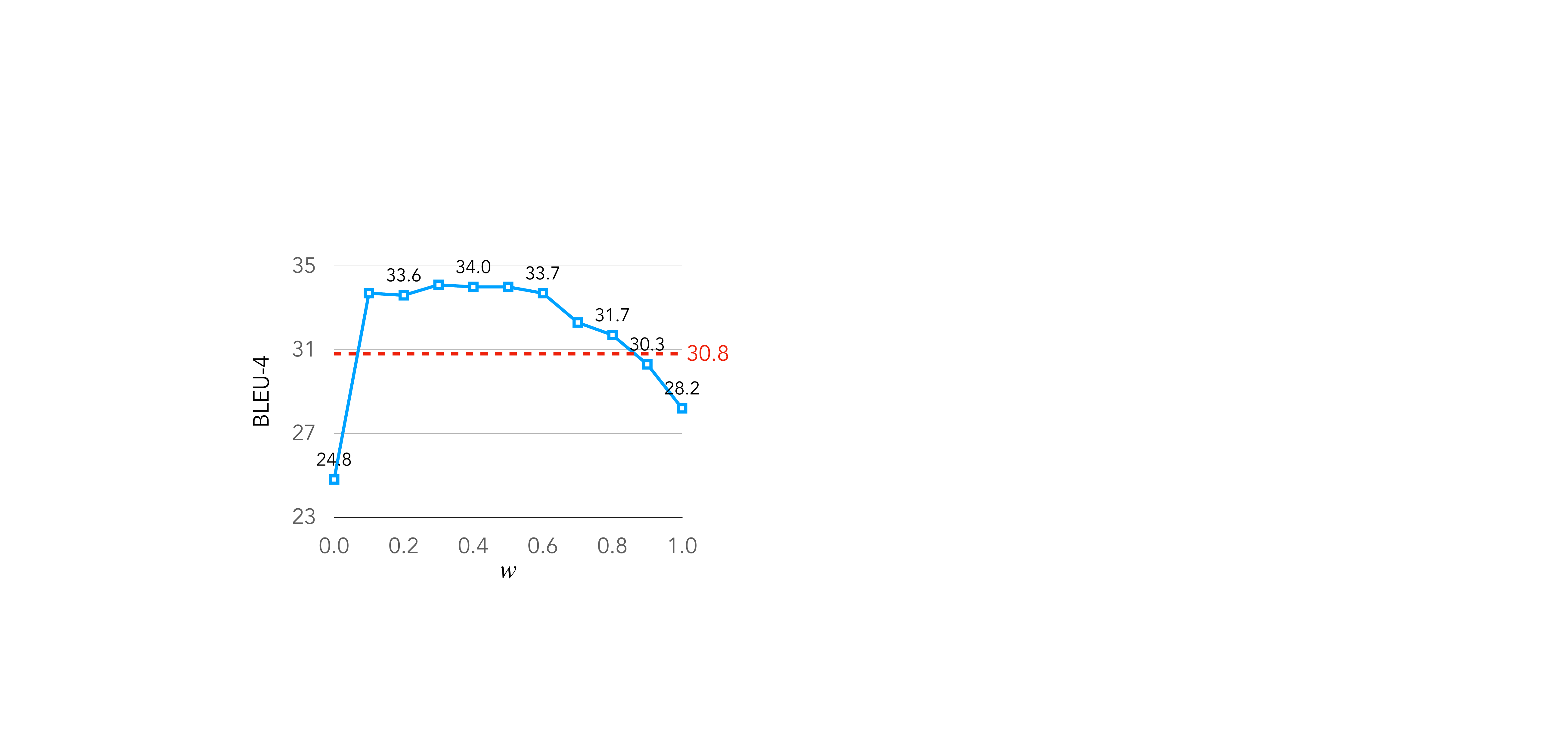}
    \caption{BLEU score on CommonGen~(dev) for different values of $w$. \ours{} achieves SoTA performance for $0.1\!\leq\!w\!\leq\!0.8$. }
    \label{fig:robustness-w}
\end{figure}

\textbf{Robustness of Hyperparameter $\boldsymbol{w}$.} As described in Section~\ref{sec:auto-generation}, for the supervised setting, the formulation of \ours{} involves a hyperparameter $0\!\leq\!w\!\leq\!1$ that decides how much the TPM or the base model contributes to generation. For our experiments, $w$ is set to 0.3 based on cross-validation results on the training set. Figure~\ref{fig:robustness-w} shows the BLEU score~(after re-ranking) on the validation set of CommonGen given different values of $w$. The performance of \ours{} is very robust with respect to different choices of $w$, achieving SoTA BLEU scores for $0.1\!\leq\!w\!\leq\!0.8$.

\begin{figure}[t]
    \centering
    \footnotesize
    \includegraphics[width=0.75\columnwidth]{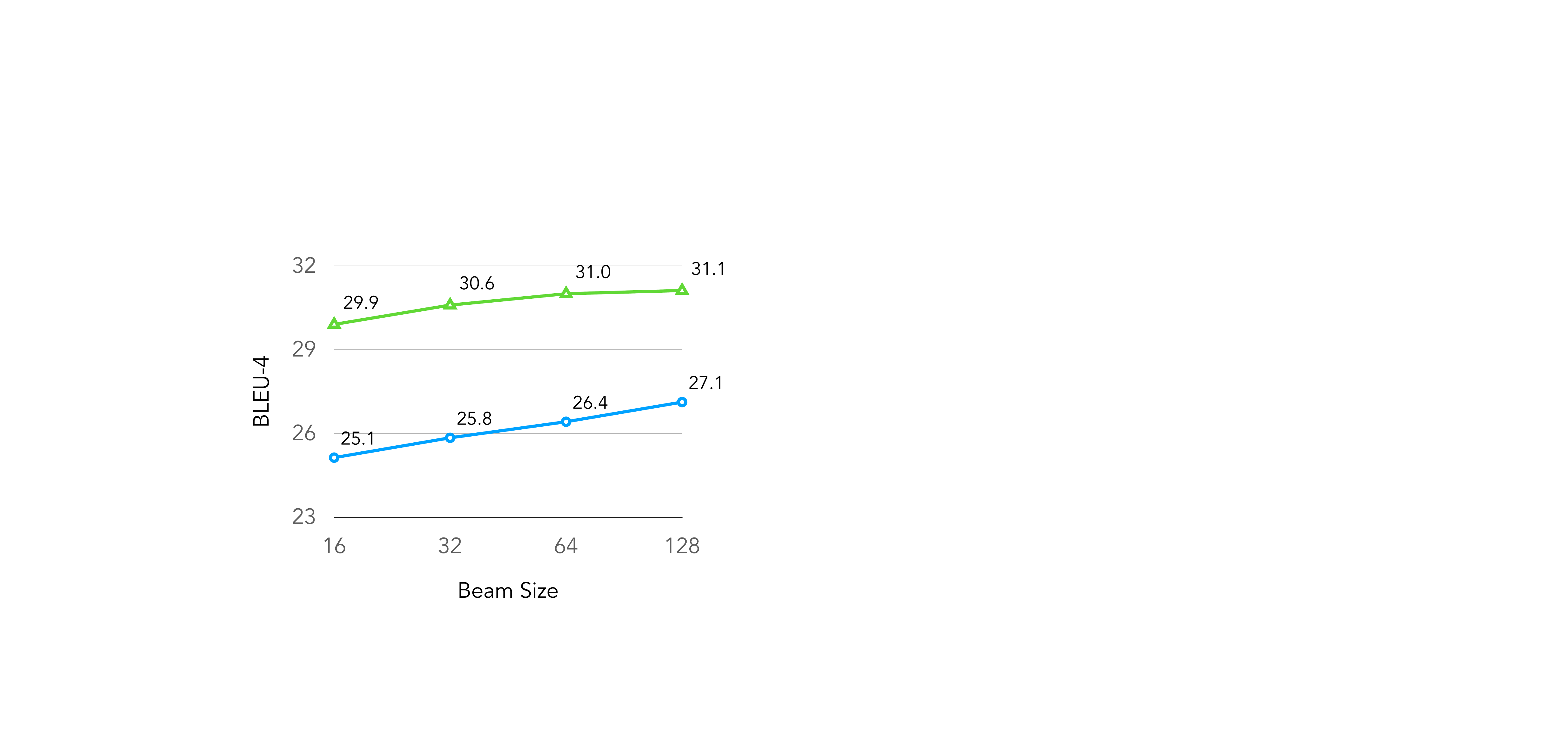}
    \vspace{-0.48em}
    \caption{\label{fig:beam-vs-bleu}
    BLEU score~(y-axis) obtained by \ours{} on CommonGen~(dev), with various beam-sizes~(x-axis), for both unsupervised~(circles) and supervised~(triangles) settings.
    }
\end{figure}

\textbf{Effect of Beam Size.} \ours{} uses beam search for generation and we study how its performance is affected by the choice of beam size. Figure~\ref{fig:beam-vs-bleu} shows that for both unsupervised and supervised settings, the performance of \ours{} improves monotonically as the beam size increases.

\textbf{Run-time Comparison.} We conduct an empirical evaluation of the run-time (in seconds) of \ours{} on CommonGen, in comparison to NeuroLogic A*esque and vanilla GPT2-large; all methods are evaluated on a single NVIDIA A100 GPU with 40 GB memory; the run-time is measured on 100 randomly sampled examples for each \# of concepts.

\ours{} achieves its best performance with beam-size=128; yet we also report the run-time for beam-size=16, where it achieves performance better than all baselines. For the unsupervised setting, \ours{} is much faster than NeuroLogic A*esque, which suffers from an unconstrained search space. For the supervised setting, \ours{} is slower than A*esque but the run-time for beam-size = 16 is still comparable.

\begin{table}[h]
    \centering
    \footnotesize
\begin{tabular}{l || c  c  c}
\toprule
\# of concepts & 3 & 4 & 5 \\
\multicolumn{1}{l}{\cellcolor{gray!25} \textit{Unsupervised}} & \multicolumn{3}{c}{\cellcolor{gray!25}} \\
A*esque & 472.9 & 542.5   & 613.9 \\ 
\ours~(16) &13.5 $\pm$ 4.4 & 21.9 $\pm$ 5.37 & 39.3 $\pm$ 6.3 \\
\ours~(128) &69.8 $\pm$ 32.3  & 97.9 $\pm$ 39.5 & 143.0 $\pm$ 44.4  \\
\multicolumn{1}{l}{\cellcolor{gray!25} \textit{Supervised}} & \multicolumn{3}{c}{\cellcolor{gray!25}} \\
A*esque & 8.5 & 9.6 & 11.4 \\ 
GPT2~(16) & 5.8 $\pm$ 1.1 & 13.0 $\pm$ 1.6 & 29.3 $\pm$ 3.2 \\
GPT2~(128) & 9.4 $\pm$ 1.8 & 21.1 $\pm$ 11.9 & 33.7 $\pm$ 3.5 \\
\ours~(16) & 11.1 $\pm$ 2.8 & 22.0 $\pm$ 5.0 & 41.6 $\pm$ 5.6 \\
\ours~(128) & 49.8 $\pm$ 20.8 & 88.7 $\pm$ 30.5 & 127.6 $\pm$ 30.4 \\
\bottomrule
\end{tabular}
    \caption{
    Time of generating one example (seconds) on CommonGen~(dev). Results for NeuroLogic A*esque, finetuned GPT2-large and \ours{} are reported; beam-sizes are shown in parentheses.}
    \label{tab:run-time}
\end{table}

\textbf{Human Evaluation.} We conduct human evaluation for sentences generated on CommonGen~(dev), following the setup of prior works~\cite{lu2022astar, meng2022nado}. Specifically, we mix sentences generated by different methods, and each sentence is presented to one human annotator to be evaluated on four aspects: concepts, plausibility, quality, and overall rating. The results are shown in Table~\ref{tab:human-evaluation}. To test statistical significance, we conduct the Wilcoxon signed rank two-sided test with $p$-value $<$ 0.05 and \ours{} performs best in all metrics compared to prior SoTA. We refer readers to Appendix~\ref{sec:human-evaluation} for details of human evaluation.
\begin{table}[th]
    \centering
    {\footnotesize
    \begin{tabular}{l||cccc}\toprule
Method      & Concepts  & Plausibility  & Quality   & Overall \\\midrule
GPT2   & 2.47  & \textbf{2.52}  & 2.65 & 2.28 \\
NADO    & \textbf{2.71}  & \textbf{2.54}  & \textbf{2.73} & 2.54 \\
\ours{} & \textbf{2.73}  & \textbf{2.52}  & \textbf{2.70} & \textbf{2.60} \\
\bottomrule
    \end{tabular}
    \caption{Human evaluation results on CommonGen for finetuned GPT2-Large, NADO and GeLaTo, all under the supervised setting.}
    \label{tab:human-evaluation}
    \vspace{-0.5em}
    }
\end{table}

\begin{table}[h]
    \centering
    {\footnotesize
    \begin{tabular}{l||cc}\toprule
\multicolumn{1}{c||}{\multirow{1}{*}{Method \textbackslash Dataset}}
& \multicolumn{1}{c}{Yelp!Review} & \multicolumn{1}{c}{News}        \\\midrule
InsNet                & 5.8             & 5.0          \\
NADO                    & 6.0             & 4.5               \\
\ours{}                 & \textbf{6.6}     &\textbf{5.4}                            \\\bottomrule
    \end{tabular}
    \caption{BLEU-4 scores for Yelp!Review and News datasets; for InsNet and NADO we present the best results of all settings while the results of \ours{} are obtained under the unsupervised setting.
    }
    }
    \label{tab:yelp_news}
\end{table}

\textbf{Fixing Order of Keywords.} Following prior works~\citep{meng2022nado,lu2022insnet}, we evaluate \ours{} on Yelp!Review and News datasets. They are more challenging in that they require keywords to appear in specific orders; besides, the average sequence lengths for both datasets are approximately 64 tokens, twice of that of CommonGen. With a minor modification to Algorithm~\ref{alg:dp}, \ours{} is easily adapted to generate text with ordered keywords. For both datasets, the training examples do not provide keywords thus there is no immediate way to finetune the base models in a supervised way. Yet, as shown in Table~\ref{tab:yelp_news}, the unsupervised \ours{} alone achieves SoTA BLEU scores.     

\section{Related Works}

\subsection{Tractable Probabilistic Models}
Tractable probabilistic models support efficient probabilistic inference~(e.g., marginal probability), thus they have been widely used in inference-demanding tasks, including enforcing algorithmic fairness~\citep{ChoiAAAI20, ChoiAAAI21}, and making predictions under missing data~\citep{khosravi2019tractable,correia2020joints,LiUAI21,DangRECOMB22}.

Probabilistic circuits (PCs) is a unified framework for a large family of tractable probabilistic models including hidden Markov models~\citep{rabiner1986introduction}, bounded tree-width graphical models~\citep{meila2000learning} and sum-product networks (SPNs)~\cite{poon2011sum}. Recent progress in learning probabilistic circuits for generative modeling~\citep{DangIJAR22,liu2022scaling} and their efficient implementation~\citep{molina2019spflow,peharz2020einsum,DangAAAI21} have been pushing the limits of PC's expressive power. 

\subsection{Enforcing Constraints in Neural Networks}
The capacity of deep generative models is continuously increasing, while their probabilistic and logic querying ability is restricted. A variety of methods have been developed. \citet{boydpredictive} introduce a general inference typology on autoregressive sequence models that can develop query estimation methods based on beam search and importance sampling. \citet{AhmedNeurIPS22} use PCs as a replacement for the SoftMax layer in neural networks such that their outputs are guaranteed to satisfy the constraint.

\subsection{Controllable Autoregressive Language Generation}
One line of research on constrained text generation focuses on modifying the decoding algorithm to inject constraints into the beam search process, such as constrained beam search~\citep{post2018fast}, NeuroLogic Decoding~\citep{lu2021neurologic} and A*esque NeuroLogic Decoding~\citep{lu2022astar}. Though they can be easily applied to various language models without training, these search-based methods can be inefficient as they suffer from large search spaces. Recent works like NADO~\citep{meng2022nado} and FUDGE~\citep{yang2021fudge} train auxiliary neural models to provide token-level guidance for autoregressive generation. Another family of approaches that enforce keyword-type constraints are insertion-based language models~\citep{lu2022insnet,susanto2020lexically}, where the initial sequences only consist of the desired keywords and the transition phrases are repeatedly inserted to complete the sentences.

\section{Conclusion}
In this paper, we propose \ours{}, where we use tractable probabilistic models~(TPMs) to impose complex lexical constraints~(denoted $\alpha$) in autoregressive language generation from large language models. Specifically, we provide token-level guidance to autoregressive generation by computing $\Pr_{\text{TPM}}(x_{t+1} \given x_{1:t}, \alpha)$. With hidden Markov model as a running example, we (1) present an efficient dynamic programming algorithm for conditioning HMMs on complex lexical constraints and (2) demonstrate the effectiveness of \ours{} on various constrained generation benchmarks; \ours{}  achieves state-of-the-art generation quality~(i.e.\ BLEU-4 scores) while guaranteeing 100\% constraint satisfaction. This work opens up new avenues for constrained language generation and motivates for the development of more expressive tractable probabilistic models.

\subsection*{Acknowledgements}
This work was funded in part by the DARPA Perceptually-enabled Task Guidance (PTG) Program under contract number HR00112220005, NSF grants \#IIS-1943641, \#IIS-1956441, \#CCF-1837129, an SRA from Meta, a research gift from Amazon Alexa AI, and a gift from RelationalAI. GVdB discloses a financial interest in RelationalAI.

\bibliography{refs}
\bibliographystyle{icml2023}

\onecolumn
\newpage
\appendix
\section{Recurrence Relation Analysis}
We establish the recurrence relation for computing $\Pr(x_{l:r},\alpha_{l:n} \given z_l)$; there are two possible cases:

Case 1. $x_{l:r} \neq \emptyset$; in this case, we can append $s \in S(x_{l:r}, \alpha)$ to $x_{l:r}$ to reduce the number of clauses in $\alpha$; abusing notation, we write $s_{i:j}$ as a shorthand for $X_{i:j} = s$:
\begin{align*}
&\Pr(\alpha_{l:n} \given z_{r+1}, x_{l:r}, z_l) \\
&\quad \!=\!\Pr(X_{r+1:r+|s|} \neq s \text{ } \forall s \in S(x_{l:r}, \alpha), \alpha_{l:n} \given  z_{r+1}, x_{l:r}, z_l) + \sum_{s \in S(x_{l:r}, \alpha)} \Pr(s_{r+1:r+|s|}, \alpha_{l:n} \given z_{r+1}, x_{l:r}, z_l) \\
&\quad \!=\!\Pr(X_{r+1:r+|s|} \neq s \text{ } \forall s \in S(x, \alpha), {\color{black} \alpha_{r+1:n}} \given  z_{r+1}) + \sum_{s \in S(x_{l:r}, \alpha)} \Pr(s_{r+1:r+|s|}, {\color{black}(\alpha \setminus x_{l:r}\oplus s)_{r+1:n}} \given z_{r+1});
\end{align*}
here $\oplus$ denotes string concatenation and $\alpha\setminus x_{l:r}\oplus s$ represents the CNF obtained by removing the clauses with any keywords appearing in $x_{l:r}\oplus s$. In particular, the second step in the derivation above follows from the non-overlapping assumption and the independence property of HMMs; then, by expanding the second term, we have:
\begin{align*}
&\Pr(\alpha_{l:n} \given z_{r+1}, x_{l:r}, z_l) \\
&\quad = \boxed{\Pr(\alpha_{r+1:n} \given z_{r+1})} \\
&\quad\quad + \sum_{s \in S(x_{l:r}, \alpha)} \boxed{\Pr(s_{r+1:r+|s|}, (\alpha \setminus x_{l:r}\oplus s)_{r+1:n} \given z_{r+1})} - \sum_{s \in S(x_{l:r}, \alpha)} \boxed{\Pr(s_{r+1:r+|s|}, \alpha_{r+1:n} \given z_{r+1})};
\end{align*}
finally, by summing over all hidden states $z_{r+1}$:
\begin{align*}
&\Pr(x_{l:r}, \alpha_{l:n} \given z_l) = \sum_{z_{r+1}} \underline{\Pr(x_{l:r},z_{r+1} \given z_l)} \Pr(\alpha_{l:n} \given z_{r+1}, x_{l:r}, z_l)
\end{align*}
Case 2. When $x = \emptyset$, we can reduce the computation of $\Pr(\alpha_{l:n} \given z_l)$ to Case 1. by summing over all possible tokens at position $l$:
\begin{align*}
&\Pr(\alpha_{l:n} \given z_{l}) \!=\! \sum_{x_l \in \text{vocabulary}} \Pr(x_l, \alpha_{l:n} \given z_{l}) \!=\! \sum_{S(x_l, \alpha) \neq \emptyset} \boxed{\Pr(x_l, \alpha_{l:n} \given z_{l})} \!+\!\sum_{S(x_l, \alpha) = \emptyset} \boxed{\Pr(x_l, \alpha_{l:n} \given z_{l})}
\end{align*}
In practice, the vocabulary size is usually large~(e.g., 50k), and most tokens lie in $\{x_l : S(x_l, \alpha) = \emptyset\}$. To avoid repetitive computation, we re-write $\sum_{S(x_l, \alpha) = \emptyset} \Pr(x_l, \alpha_{l:n} \given z_{l})$:
\begin{align*}
&{\sum}_{S(x_l, \alpha) = \emptyset} \Pr(x_l, \alpha_{l:n} \given z_{l}) \\
&\quad\!=\!{\sum}_{S(x_l, \alpha)=\emptyset}{\sum}_{z_{l+1}}\Pr(x_l, \alpha_{l:n}, z_{l+1}\given z_{l}) \\
&\quad\!=\! \left({\sum}_{S(x_l, \alpha)=\emptyset}\underline{\Pr(x_l\given z_l)}\right) \cdot \left({\sum}_{z_{l+1}}\underline{\Pr(z_{l+1} \given z_{l})} \boxed{\Pr(\alpha_{l+1:n} \given z_{l+1})} \right)
\end{align*}
where $\sum_{z_{l+1}} \Pr(z_{l+1} \given z_{l}) \Pr(\alpha_{l+1:n} \given z_{l+1})$ does not depend on $x_l$ and the summation ${\sum}_{S(x_l, \alpha)=\emptyset}\Pr(x_l\given z_l)$ can be efficiently computed with CUDA parallelization without enumerating over all tokens.

\newpage
\section{Human Evaluation Setup}
\label{sec:human-evaluation}
The following screenshot shows the human evaluation setup for CommonGen. We consider \emph{Yes} as 3 points, \emph{Somewhat} as 2 points and \emph{No} as 1 point.

\includegraphics[width=0.98\linewidth]{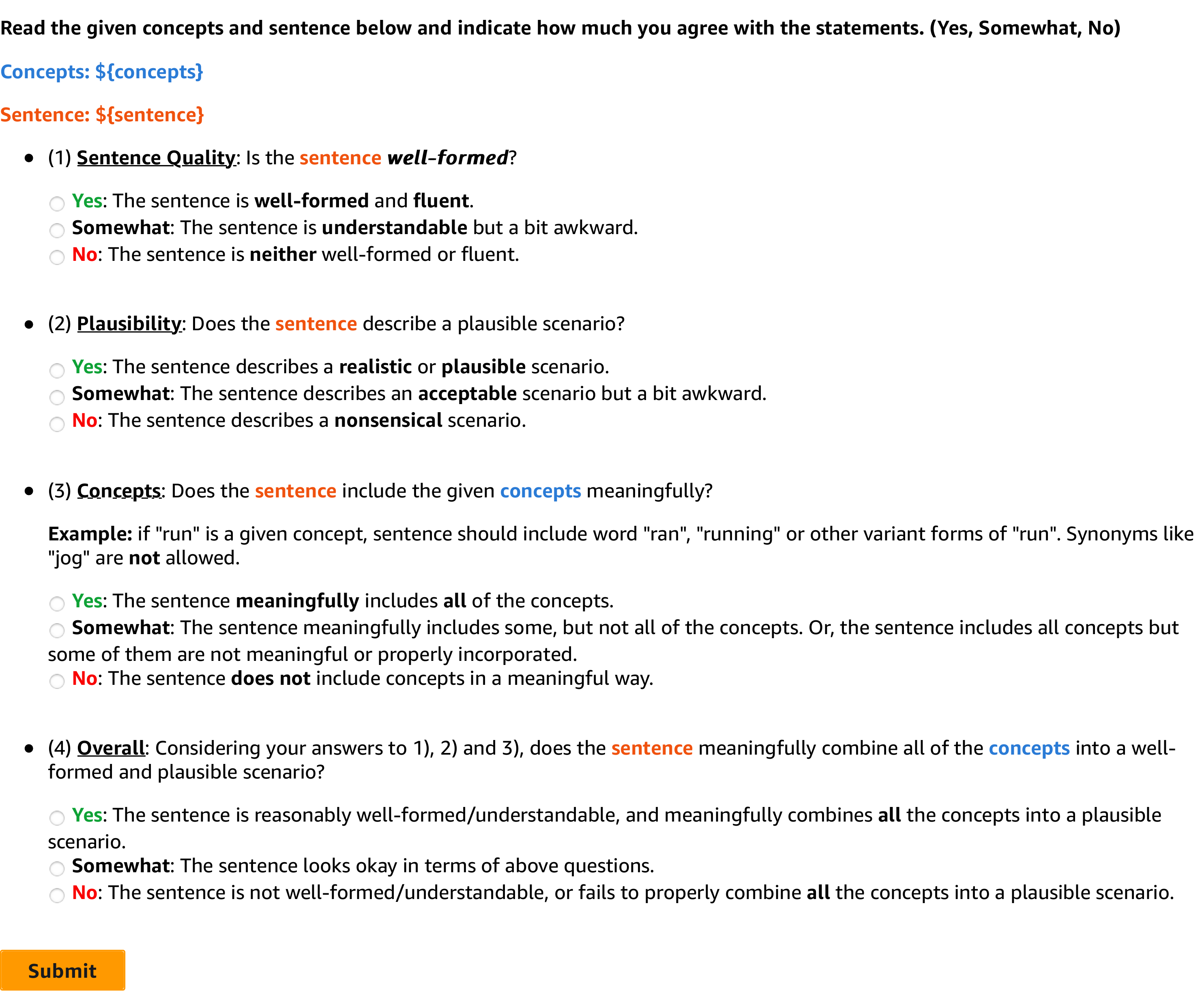}

\end{document}